# DA-SSL: SELF-SUPERVISED DOMAIN ADAPTOR TO LEVERAGE FOUNDATIONAL MODELS IN TURBT HISTOPATHOLOGY SLIDES

*Haoyue Zhang[1], Meera Chappidi[2,3], Erolcan Sayar[2], Helen Richards[2], Zhijun Chen[1], Lucas Liu[2], Roxanne Wadia[3], Peter A Humphrey[3], Fady Ghali[3], Alberto Contreras-Sanz[4], Peter Black[4], Jonathan Wright[5], Stephanie Harmon[1,#], Michael Haffner[2,#]*

[1]National Cancer Institute, National Institutes of Health, Bethesda, MD, USA
[2]Fred Hutchinson Cancer Center, Seattle, WA, USA
[2]University of California, San Francisco, San Francisco, CA, USA
[3]Yale University, New Haven, CT, USA
[4]University of British Columbia, Vancouver, BC, Canada
[5]University of Washington, Seattle, WA, USA

## ABSTRACT

Recent deep learning frameworks in histopathology, particularly multiple instance learning (MIL) combined with pathology foundational models (PFMs), have shown strong performance. However, PFMs exhibit limitations on certain cancer or specimen types due to domain shifts—these cancer types were rarely used for pretraining or specimens contain tissue-based artifacts rarely seen within the pretraining population. Such is the case for transurethral resection of bladder tumor (TURBT), which are essential for diagnosing muscle-invasive bladder cancer (MIBC), but contain fragmented tissue chips and electrocautery artifacts and were not widely used in publicly available PFMs. To address this, we propose a simple yet effective domain-adaptive self-supervised adaptor (DA-SSL) that realigns pretrained PFM features to the TURBT domain without fine-tuning the foundational model itself. We pilot this framework for predicting treatment response in TURBT, where histomorphological features are currently underutilized and identifying patients who will benefit from neoadjuvant chemotherapy (NAC) is challenging. In our multi-center study, DA-SSL achieved an AUC of 0.77 ± 0.04 in five-fold cross-validation and an external test accuracy of 0.84, sensitivity of 0.71, and specificity of 0.91 using majority voting. Our results demonstrate that lightweight domain adaptation with self-supervision can effectively enhance PFM-based MIL pipelines for clinically challenging histopathology tasks. Code is available at: https://github.com/zhanghaoyue/DA_SSL_TURBT.

## 1. INTRODUCTION

Recent pathology foundational models (PFMs) have achieved impressive generalization across tissue types. However, underlying bias is a growing concern because lack of representation [1], [2], additionally cancer specimens and artifact patterns differ greatly from the large resection or biopsy slides used in PFM pretraining. A primary example of this is Transurethral resection of bladder tumor (TURBT). TURBT slides present unique computational challenges: tissue is fragmented into small resection chips, and electrocautery often introduces artifacts that obscure morphology and complicate both staging and algorithmic analysis [3]. This motivates efficient domain adaptation strategies for pathology-only workflows. Among recent methods, the Re-embedded Regional Transformer (RRT)[4] enhances multiple-instance learning (MIL) by re-embedding regional features through transformer blocks. While effective for coherent resection slides, RRT relies on global spatial relationships that are absent in fragmented TURBT specimens. SlideGraph[5] similarly tackles TURBT-related task by integrating whole-slide and genomic data via graph modeling, but its multimodal design limits direct application to single-modality pathology.

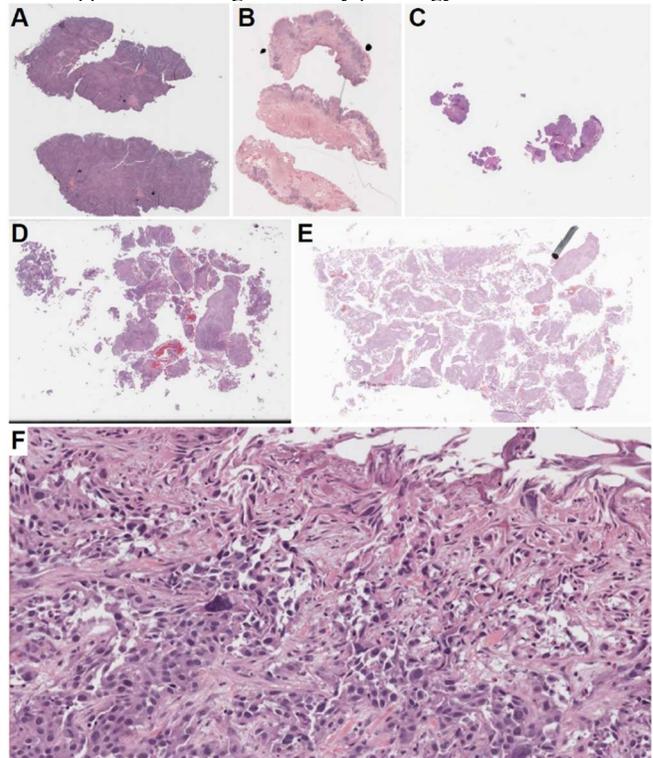

Fig. 1. Samples of two major unique characteristics of the TURBT slide. The upper section shows multiple random TURBT slides from our cohort. The specimen exhibits variation in both global distribution and morphology, ranging from large chunks to small fragments and scattered samples. The lower section displays the TURBT slide artifact, for which common artifact-detection models are ineffective.

We propose a Domain-Adaptive Self-Supervised Learning (DA-SSL) framework that realigns frozen PFM features to the TURBT domain. DA-SSL attaches a lightweight adaptor to the MIL encoder and jointly optimizes both via SimSiam self-supervision, refining morphology-specific embeddings without modifying the foundation model. Our contributions are threefold:

1. We demonstrate and quantify the domain limitations of PFMs on TURBT slides.
2. We propose an efficient adaptor enabling joint self-supervised and supervised training within the MIL framework.
3. We validate DA-SSL on multi-center cohorts, achieving improved generalization and outperforming both pathology-only and multimodal baselines.

We pilot DA-SSL for predicting treatment response from TURBT specimens. Muscle-invasive bladder cancer (MIBC), which requires aggressive multimodal therapy [6][7], is diagnosed from TURBT prior to treatment decision-making [8]. Neoadjuvant chemotherapy (NAC) followed by radical cystectomy (RC) is a first-line treatment for MIBC and improves survival compared with RC alone[9] but only ~30% of patients achieve a complete pathologic response (pCR), and 20–30% experience disease progression during NAC[10], [11], [12], [13], highlighting the need for accurate pretreatment response prediction.

## 2. METHOD

### 2.1. TURBT-Aware Preprocessing

First, a **Tumor and Artifact Filter** is used. Whole-slide images (WSIs) were processed using Trident[14] to generate patch-level features and tissue-type predictions (tumor, normal, artifact). To remove electrocautery artifacts and non-diagnostic regions, we trained a 3-class ViT[15] classifier initialized with UNI-v1[16] weights. The model was first trained on TCGA bladder cystectomy slides (tumor vs. normal) with spleen and lymph node samples as negative controls, then fine-tuned on 40 expert-annotated TURBT slides across three refinement rounds. During inference, only tumor patches were retained.

Second, **Uniform Spatial Sampling** is used to stabilize bag size and spatial coverage across fragmented chips. Patch coordinates were normalized to $[0,1]^2$ and discretized into a $G \times G$ Grid. Patches were uniformly sampled from each grid until a maximum of K patches was reached. Only during self-supervised training, bags were zero-padded to length K and accompanied by a Boolean mask for batched processing.

### 2.2. Foundation Model Feature Extraction

Patch embeddings were extracted from pathology foundation models (PFMs): UNI v1/v2, Virchow v2[17], Prov-GigaPath[18], and H-Optimus-1[19]. This aligns with standard MIL practice where downstream modules operate on frozen features. Extraction was performed using in-house code, along with tumor labels from the classifier were stored with coordinates.

### 2.3. Self-Supervised Domain Adapter

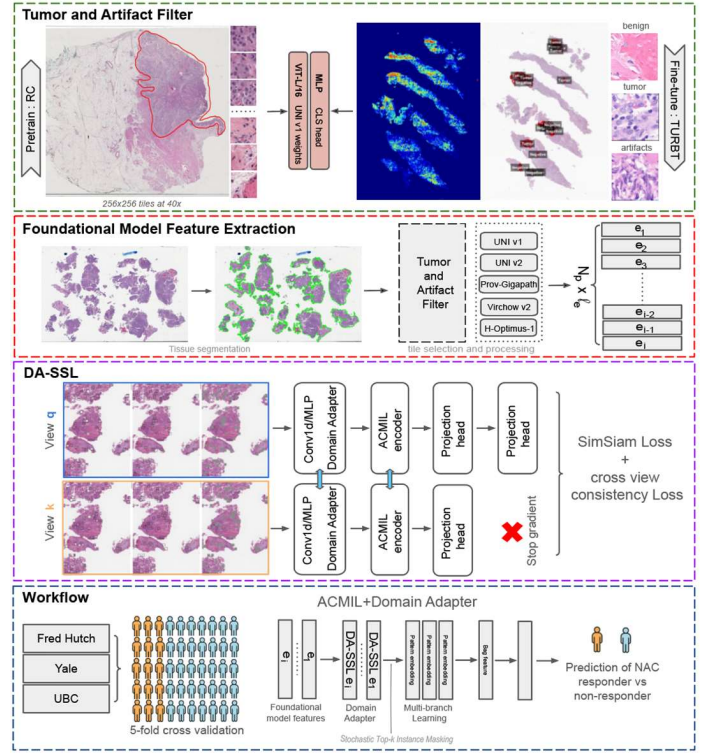

Fig. 2. Framework Overview.

DA-SSL adapts frozen PFM embeddings to the TURBT domain using a SimSiam[20] objective aligned to MIL representations.

**Adapter Design.** Given feature embedding $x \in R^{B \times K \times d}$, where B is the batch size, K is the number of patches, and d is the feature embedding length,
- MLPFeatureAdapter: $z = x + MLP(x)$ with two linear layers and ReLU in between.
- Conv1DFeatureAdapter: $z = x + Conv1D(x)$ with two conv1d layers and ReLU in between.

Both designs aim for minimal complexity and a residual connection to help stabilize the feature updates.

**MIL-space projection and predictor.** A 3-layer projector (Linear–ReLU–InstanceNorm1D) maps features to the SimSiam latent space, followed by a predictor head (InstanceNorm1D–ReLU–Linear).

**Forward and Loss.** For two independently augmented views $(x^{(1)}, x^{(2)})$ from the same bag,
$$z_1 = Adapter(x^{(1)}), z_2 = Adapter(x^{(2)}),$$
$$(\cdot, m_1) = MIL(z_1, mask), (\cdot, m_2) = MIL(z_2, mask),$$
$$p_1 = Pred(m_1), p_2 = Pred(m_2),$$
We minimize the standard SimSiam loss:
$$L_{SimSiam} = \frac{1}{2}(-cos\,(p_1, sg(m_2)) - cos\,(p_2, sg(m_1)),$$
Where sg is stop-gradient on the target branch, only the adapter+projector+predictor is updated during the DA-SSL process; the FM features are just treated the same way as the input image. To stabilize early training, we add a cross-view consistency term that encourages patch-level invariance:
$$L_{cv-cons} = E_M ||z_1 - z_2||_2^2 + \lambda E_M ||z_1||_1$$
Where $\lambda$ is 0.1. Therefore, our final loss is:
$$L_{DA-SSL} = L_{SimSiam} + 0.5 \times L_{cv-c}$$

TABLE I
NAC RESPONSE PREDICTION PERFORMANCE

| PFM | Methods | Multi-Center Cross-Validation | | | | | |
|---|---|---|---|---|---|---|---|
| | | AUC | Accuracy | Sensitivity | Specificity | Precision | F1-score |
| Res-Net50 | ABMIL | 57.95±2.73 | 61.42±4.40 | 7.25±12.16 | 91.68±9.35 | 9.22±11.98 | 7.60±11.74 |
| | ACMIL | 60.73±5.66 | 58.28±10.70 | 76.90±19.35 | 47.21±21.91 | 45.63±7.61 | 55.23±7.39 |
| | ACMIL+RRT | 64.59±3.61 | 61.53±3.77 | 48.83±23.05 | 71.21±13.25 | 44.87±10.06 | 42.45±16.08 |
| UNI v1 | ACMIL | 73.37±3.79 | 67.59±5.58 | 69.16±1828 | 66.69±18.32 | 55.66±14.02 | 58.31±5.94 |
| | ACMIL+RRT | 70.34±3.10 | 70.60±4.16 | 65.91±11.66 | 71.63±11.36 | 56.37±5.62 | 59.96±5.01 |
| | +DA-MLP | 74.71±3.99 | 67.65±6.74 | 72.75±14.65 | 66.76±19.29 | 56.72±16.49 | 60.18±4.84 |
| | +DA-MLP+SSL | 75.00±2.15 | 70.25±3.53 | 69.96±15.00 | 69.81±14.97 | 56.26±9.85 | 60.81±5.92 |
| | +DA-Conv1d | 73.69±4.90 | 73.41±6.57 | 71.65±9.97 | 73.99±11.23 | 60.54±5.30 | 64.92±3.06 |
| | +DA-Conv1d+SSL | 74.09±1.48 | 70.03±1.96 | 74.00±9.41 | 67.86±7.59 | 54.81±8.62 | 62.13±6.12 |
| UNI v2 | ACMIL | 73.78±5.70 | 70.42±6.48 | 67.57±8.51 | 71.72±9.65 | 56.79±9.80 | 60.93±7.46 |
| | +DA-MLP | 74.35±3.03 | 71.75±3.31 | 74.84±5.83 | 69.42±5.23 | 56.05±7.69 | 63.90±6.72 |
| | +DA-MLP+SSL | 75.85±3.19 | 72.27±5.05 | 75.85±10.55 | 68.13±12.56 | 57.40±1.63 | 64.91±3.86 |
| | +DA-Conv1d | 74.29±3.35 | 71.79±73.78 | 73.38±11.05 | 69.77±11.56 | 57.92±6.30 | 63.93±3.59 |
| | +DA-Conv1d+SSL | 74.72±2.09 | 68.87±2.76 | 75.59±9.79 | 66.14±7.99 | 54.04±9.81 | 61.78±5.92 |
| Prov-Gigapth | ACMIL | 72.42±7.98 | 68.03±7.88 | 74.95±8.80 | 64.23±14.64 | 60.12±11.63 | 65.54±6.90 |
| | +DA-MLP | 76.22±2.02 | 72.23±2.66 | 68.75±8.84 | 73.25±7.62 | 57.73±8.00 | 62.26±6.15 |
| | +DA-MLP+SSL | <u>77.28±3.83</u> | <u>75.13±4.85</u> | 66.91±13.75 | <u>79.68±10.34</u> | <u>64.44±11.28</u> | <u>63.93±7.89</u> |
| | +DA-Conv1d | 76.86±5.72 | 73.89±3.46 | 63.98±12.00 | 79.22±4.54 | 61.44±4.81 | 61.88±5.90 |
| | +DA-Conv1d+SSL | 77.03±4.12 | 72.90±6.18 | 76.67±7.63 | 71.30±13.15 | 59.68±12.90 | 65.80±6.53 |
| Virchow v2 | ACMIL | 72.92±7.08 | 70.57±5.79 | 64.49±13.15 | 76.16±11.96 | 62.14±14.94 | 60.31±3.75 |
| | +DA-MLP | 74.52±6.62 | 71.23±5.01 | 58.78±16.13 | 78.51±11.65 | 60.30±10.85 | 57.25±7.25 |
| | +DA-MLP+SSL | 76.57±6.06 | 70.82±3.46 | 51.34±28.70 | 79.78±47.62 | 47.62±25.29 | 48.06±24.55 |
| | +DA-Conv1d | 73.76±2.38 | 68.04±3.37 | 79.07±62.86 | 62.86±13.33 | 54.35±11.69 | 62.00±5.50 |
| | +DA-Conv1d+SSL | 75.47±6.00 | 72.61±5.42 | 69.65±12.57 | 73.28±9.88 | 58.97±7.07 | 62.96±6.39 |
| H-Optimus 1 | ACMIL | 71.94±3.74 | 67.12±10.32 | 74.13±13.29 | 65.05±21.00 | 57.59±15.85 | 61.61±6.49 |
| | +DA-MLP | 74.63±6.18 | 69.53±6.45 | 70.76±15.58 | 71.34±13.83 | 59.35±12.30 | 61.60±3.85 |
| | +DA-MLP+SSL | 76.72±7.09 | 70.98±4.45 | 81.25±10.46 | 66.36±9.59 | 56.20±10.13 | 65.28±5.11 |
| | +DA-Conv1d | 76.38±6.18 | 71.13±5.43 | 76.16±7.12 | 68.34±7.96 | 55.88±9.37 | 63.98±6.92 |
| | +DA-Conv1d+SSL | **77.35±4.36** | **71.47±6.04** | **79.50±14.38** | **67.95±16.94** | **59.74±15.96** | **65.08±5.64** |

Mean ± std from 5-fold cross-validation experiments. The best model is highlighted in bold, and the second best is highlighted in underline.

### 2.4. View Generation and Loader for DA-SSL

We implement a slide-level data loader to produce two feature-space views per slide:
1. **Load & filter**: read feature file (HDF5) with features, coordinates, and tumor classification labels. Retain tumor regions and drop rows with NaN values.
2. **Grid sample**: Apply the 32x32 uniform sampling to obtain up to K tokens; pad with zeros to length K; shuffle rows.
3. **Two views generation**: apply the following instance-level augmentation chain twice and independently to the same padded feature matrix, inspired by [21]:
   **Instance Masking** – mask entire instance vectors
   **Instance feature Replace** – replaces a subspace of features from another instance in the bag
   **Instance Replace** – swaps a full instance vector with another instance within the same bag
   **Instance Feature Noise** – adds small Gaussian Noise
   **Instance Feature Drop** – drops a contiguous feature block
   **Instance Feature Dropout** – per-dimension random dropout

### 2.5. Slide-Level Classifier (MIL)

We adopt ACMIL, a multi-branch attention MIL model, for slide-level prediction. ACMIL extends ABMIL with multiple attention heads and masking regularization to encourage broader patch exploration. Baselines included ABMIL with ResNet-50 features to contextualize task difficulty.

## 3. EXPERIMENTS AND RESULTS

### 3.1. Dataset

This multicenter study included 355 TURBT slides from 249 MIBC patients across four institutions: Fred Hutch/UW (n=160), Yale (n=71), UBC (n=124). Eligible cases received ddMVAC, MVAC, or Gem/Cis chemotherapy regimens. An additional 311 WSIs (296 TCGA-BLCA cystectomies + 15 spleen/lymph node slides) were used only for pretraining the tumor–artifact filter. The responder vs non-responder ratio is 35:65. All slides were quality-checked by a pathologist and refined with GrandQC segmentation or traditional Otsu's thresholding, with patch extraction at 20× (256×256 px) and normalization per foundation model (PFM) requirements.

### 3.2. Results

ResNet-50 ImageNet feature embeddings + ABMIL achieved an AUC of 57.9±2.7, consistent with previous reports [5] and highlighting the difficulty of modeling TURBT morphology. Substituting ACMIL improved AUC to 60.7 ± 5.7 and feature re-embedding (RRT) further raised it to 64.6 ± 3.6. However, when applied to PFM features, RRT degraded performance, likely because transformer layers are ill-suited to the fragmented spatial structure of TURBT slides with its reliance on global self-attention. Direct ACMIL training with five PFMs (UNI v1/v2, Virchow v2, Prov-GigaPath, H-Optimus-1) achieved 71.9–73.8 AUC, clearly outperforming ImageNet-based baselines with a 14-16-point improvement. These results confirm that pretrained pathology embeddings substantially enhance downstream prediction. Applying the proposed Domain-Adaptive Self-Supervised Learning (DA-SSL) framework led to consistent gains for all PFMs. Prov-GigaPath and H-Optimus-1 achieved the largest relative improvements, while UNI v1/v2 saw smaller but steady boosts. DA-SSL also reduced fold-to-fold variance, indicating more stable convergence and improved feature invariance. These results confirm that lightweight feature-space adaptation enhances foundational representations for domain-specific histopathology tasks (Table I). Our models outperform prior reports [5], which achieved 69.38±5.65 / 67.27 in cross-validation/test with imaging only and 74.00±10.00 / 72.00 cross-validation/test with multi-modal approach.

### 3.3. Ablation Study

When comparing MIL backbone variants using the same UNI v1 features (Table II), ABMIL achieved 72.8 ± 4.6 AUC, outperforming CLAM (56.2 ± 11.8), TransMIL (64.8 ± 5.3), and DS-MIL (65.9 ± 5.3). Methods that rely on instance-level pseudo-labeling or long-range attention failed to generalize to the small, discontinuous tissue fragments characteristic of TURBT. In contrast, simpler ABMIL and ACMIL effectively aggregate weak, distributed cues, proved to be more suited for TURBT slides. Uniform grid sampling slightly reduced mean AUC but stabilized training, while the tumor–artifact filter improved focus on diagnostic regions and reduced noise from non-tumor tissue (Table III).

### 4. DISCUSSION AND CONCLUSION

Predicting neoadjuvant chemotherapy (NAC) response in muscle-invasive bladder cancer (MIBC) directly from transurethral resection of bladder tumor (TURBT) slides is both clinically meaningful and technically demanding. TURBT specimens are inherently fragmented, frequently distorted by cautery, and exhibit heterogeneous tumors and stromal composition, producing weak and spatially diffuse histologic signals. While recent pathology foundational models (PFMs) demonstrate strong generalization across tissue types, our study shows that their pretrained embeddings alone are insufficient for this highly domain-specific setting. The features extracted from PFMs provide a strong starting point—enabling a level of predictive performance previously unattainable with conventional CNN features—but their fixed representations quickly reach a performance ceiling. Without an effective strategy to adapt or refine these frozen embeddings, shallow multiple-instance learning (MIL) architectures can only marginally improve downstream accuracy.

The proposed DA-SSL framework effectively bridges this gap by attaching a lightweight adaptor to frozen PFM features and jointly training it with the MIL encoder under a SimSiam objective. This strategy enables domain-specific representation refinement without costly model fine-tuning, outperforming RRT as methodological baseline and SlideGraph+ on the same NAC-response prediction benchmark [5]. DA-SSL also stabilizes training under weak, spatially scattered signals typical of TURBT morphology. Although the cohort size remains modest, our findings highlight the potential of efficient, self-supervised domain adaptation to extend foundational pathology models toward challenging specimen types and clinically actionable prediction tasks.

TABLE II
Comparison of different MIL frameworks

| MIL | AUC | ACC | SENS | SPEC | PREC | F1 |
|---|---|---|---|---|---|---|
| AB-MIL | 72.76 ±3.63 | 68.36 ±4.58 | 72.66± 14.3 | 67.14± 9.40 | 53.99±10.99 | 60.08 ±8.12 |
| CLAM | 56.21 ±11.8 | 58.37 ±13.4 | 56.95± 26.9 | 57.64± 24.88 | 45.25±10.35 | 45.35 ±16.2 |
| Trans-MIL | 64.78 ±5.32 | 60.06 ±6.65 | 73.81± 10.6 | 53.17± 11.83 | 45.68±11.33 | 55.37 ±9.40 |
| DS-MIL | 65.94 ±5.25 | 63.07 ±10.1 | 62.08± 23.3 | 63.38± 22.50 | 52.78±1.84 | 51.91 ±8.81 |
| AC-MIL | 73.37 ±3.79 | 67.59 ±5.58 | 69.16± 18.3 | 66.69± 18.32 | 55.66±14.02 | 58.31 ±5.94 |

Investigate the impact of different MIL frameworks with UNI v1 feature embeddings. The performance is based on the development dataset and 5-fold cross-validation. ACC=Accuracy, SENS=Sensitivity, SPEC=Specificity, PREC=Precision

TABLE III
Impact of tumor and artifact filter; uniform grid sampling

| | AUC | ACC | SENS | SPEC | PREC | F1 |
|---|---|---|---|---|---|---|
| A | 70.25± 10.16 | 66.22± 9.55 | 65.39± 12.76 | 68.42± 15.16 | 60.42± 12.28 | 60.99± 8.46 |
| B | 69.92± 1.56 | 66.60± 4.38 | 71.73± 10.34 | 61.92± 11.41 | 55.44± 8.21 | 61.84± 7.03 |
| C | 73.37± 3.79 | 67.59± 5.58 | 69.16± 18.28 | 66.69± 18.32 | 55.66± 14.02 | 58.31± 5.94 |

A = ACMIL alone, B = ACMIL + uniform grid sampling, C = ACMIL + uniform grid sampling + tumor-artifact filter. ACC=Accuracy, SENS=Sensitivity, SPEC=Specificity, PREC=Precision

### 4. COMPLIANCE WITH ETHICAL STANDARDS

This study was completed following Institutional Review Board approval at each of the following universities: the University of Washington, Yale University, and the University of British Columbia.

### 5. ACKNOWLEDGEMENT

This research was supported in part by the Center for Cancer Research, National Cancer Institute, National Institutes of Health Intramural Research Program project number ZIABC012163. The findings and conclusions presented in this paper are those of the author(s) and do not necessarily reflect the views of the NIH or the U.S. Department of Health and Human Services.

## 6. REFERENCES

[1] A. Vaidya *et al.*, "Demographic bias in misdiagnosis by computational pathology models.," *Nat Med*, vol. 30, no. 4, pp. 1174–1190, Apr. 2024, doi: 10.1038/s41591-024-02885-z.

[2] E. D. de Jong, E. Marcus, and J. Teuwen, "Current pathology foundation models are unrobust to medical center differences," *arXiv preprint arXiv:2501.18055*, 2025.

[3] M. Truong, L. Liang, J. Kukreja, J. O'Brien, J. Jean-Gilles, and E. Messing, "Cautery artifact understages urothelial cancer at initial transurethral resection of large bladder tumours," *CUAJ*, vol. 11, no. 5, p. 203, May 2017, doi: 10.5489/cuaj.4172.

[4] W. Tang, F. Zhou, S. Huang, X. Zhu, Y. Zhang, and B. Liu, "Feature Re-Embedding: Towards Foundation Model-Level Performance in Computational Pathology," Jul. 25, 2024, *arXiv*: arXiv:2402.17228. doi: 10.48550/arXiv.2402.17228.

[5] Z. Bai *et al.*, "Predicting response to neoadjuvant chemotherapy in muscle-invasive bladder cancer via interpretable multimodal deep learning," *npj Digit. Med.*, vol. 8, no. 1, p. 174, Mar. 2025, doi: 10.1038/s41746-025-01560-y.

[6] R. L. Siegel, K. D. Miller, N. S. Wagle, and A. Jemal, "Cancer statistics, 2023," *CA A Cancer J Clinicians*, vol. 73, no. 1, pp. 17–48, Jan. 2023, doi: 10.3322/caac.21763.

[7] L. Dyrskjøt *et al.*, "Bladder cancer," *Nat Rev Dis Primers*, vol. 9, no. 1, p. 58, Oct. 2023, doi: 10.1038/s41572-023-00468-9.

[8] L. H. C. Kim and M. I. Patel, "Transurethral resection of bladder tumour (TURBT)," *Transl Androl Urol*, vol. 9, no. 6, pp. 3056–3072, Dec. 2020, doi: 10.21037/tau.2019.09.38.

[9] Advanced Bladder Cancer Meta-analysis Collaboration, "Neo-adjuvant chemotherapy for invasive bladder cancer," *Cochrane Database of Systematic Reviews*, vol. 2012, no. 2, Jan. 2004, doi: 10.1002/14651858.CD005246.

[10] H. B. Grossman *et al.*, "Neoadjuvant Chemotherapy plus Cystectomy Compared with Cystectomy Alone for Locally Advanced Bladder Cancer," *N Engl J Med*, vol. 349, no. 9, pp. 859–866, Aug. 2003, doi: 10.1056/NEJMoa022148.

[11] International Collaboration of Trialists on behalf of the Medical Research Council Advanced Bladder Cancer Working Party, "International Phase III Trial Assessing Neoadjuvant Cisplatin, Methotrexate, and Vinblastine Chemotherapy for Muscle-Invasive Bladder Cancer: Long-Term Results of the BA06 30894 Trial," *JCO*, vol. 29, no. 16, pp. 2171–2177, Jun. 2011, doi: 10.1200/JCO.2010.32.3139.

[12] M. Mossanen *et al.*, "Nonresponse to Neoadjuvant Chemotherapy for Muscle-Invasive Urothelial Cell Carcinoma of the Bladder," *Clinical Genitourinary Cancer*, vol. 12, no. 3, pp. 210–213, Jun. 2014, doi: 10.1016/j.clgc.2013.10.002.

[13] H. Zargar *et al.*, "Multicenter Assessment of Neoadjuvant Chemotherapy for Muscle-invasive Bladder Cancer," *European Urology*, vol. 67, no. 2, pp. 241–249, Feb. 2015, doi: 10.1016/j.eururo.2014.09.007.

[14] A. Zhang, G. Jaume, A. Vaidya, T. Ding, and F. Mahmood, "Accelerating Data Processing and Benchmarking of AI Models for Pathology," Feb. 10, 2025, *arXiv*: arXiv:2502.06750. doi: 10.48550/arXiv.2502.06750.

[15] A. Dosovitskiy *et al.*, "An Image is Worth 16x16 Words: Transformers for Image Recognition at Scale," Jun. 03, 2021, *arXiv*: arXiv:2010.11929. doi: 10.48550/arXiv.2010.11929.

[16] R. J. Chen *et al.*, "Towards a general-purpose foundation model for computational pathology," *Nat Med*, vol. 30, no. 3, pp. 850–862, Mar. 2024, doi: 10.1038/s41591-024-02857-3.

[17] E. Zimmermann *et al.*, "Virchow2: Scaling Self-Supervised Mixed Magnification Models in Pathology," Nov. 06, 2024, *arXiv*: arXiv:2408.00738. doi: 10.48550/arXiv.2408.00738.

[18] H. Xu *et al.*, "A whole-slide foundation model for digital pathology from real-world data," *Nature*, vol. 630, no. 8015, pp. 181–188, Jun. 2024, doi: 10.1038/s41586-024-07441-w.

[19] Bioptimus, *H-optimus-1*. (2025). [Online]. Available: https://huggingface.co/bioptimus/H-optimus-1

[20] X. Chen and K. He, "Exploring Simple Siamese Representation Learning," 2020, *arXiv*. doi: 10.48550/ARXIV.2011.10566.

[21] Y. Wang, Y. Zhao, Z. Wang, and M. Wang, "Robust Self-Supervised Multi-Instance Learning with Structure Awareness," *AAAI*, vol. 37, no. 8, pp. 10218–10225, Jun. 2023, doi: 10.1609/aaai.v37i8.26217.